\documentclass[conference]{IEEEtran}
\usepackage{array}

\usepackage{xcolor}

\usepackage{cite}

\ifCLASSINFOpdf
  \usepackage[pdftex]{graphicx}
  \graphicspath{{../figures/}}
  \DeclareGraphicsExtensions{.pdf,.jpeg,.png}
\else
  \usepackage[dvips]{graphicx}
  \graphicspath{{../figures/}}
  \DeclareGraphicsExtensions{.eps}
\fi
\usepackage{tikz}
\usetikzlibrary{patterns}
\usepackage{pgfplots}
\usepackage{siunitx}
\usepgfplotslibrary{units} 
\usetikzlibrary{pgfplots.groupplots}

\definecolor{spprl-red}{HTML}{d7191c}
\definecolor{spprl-light-orange}{HTML}{fdae61}
\definecolor{spprl-light-blue}{HTML}{abd9e9}
\definecolor{spprl-blue}{HTML}{2c7bb6}

\usepackage{balance}

\usepackage{amsmath}
\usepackage{algpseudocode,algorithm,algorithmicx}

\makeatletter
\newcommand\fs@norules{\def\@fs@cfont{\bfseries}\let\@fs@capt\floatc@ruled
  \def\@fs@pre{}%
  \def\@fs@post{}%
  \def\@fs@mid{\kern3pt}%
  \let\@fs@iftopcapt\iftrue}
\makeatother
\floatstyle{norules}
\restylefloat{algorithm}
\ifCLASSOPTIONcompsoc
  \usepackage[caption=false,font=normalsize,labelfont=sf,textfont=sf]{subfig}
\else
  \usepackage[caption=false,font=footnotesize]{subfig}
\fi
\makeatletter
\newcommand*{\centerfloat}{%
  \parindent \z@
  \leftskip \z@ \@plus 1fil \@minus \textwidth
  \rightskip\leftskip
  \parfillskip \z@skip}
\makeatother

\usepackage{tabularx}
\usepackage{adjustbox}
\usepackage{multirow}

\usepackage{url}

\usepackage[margins]{trackchanges} 

\usepackage{soul} 

\usepackage[normalem]{ulem} 

\usepackage{bm} 

\begin{document}
\bstctlcite{MyBSTcontrol}

\title{Attack Strength vs. Detectability Dilemma in Adversarial Machine Learning}

\author{\IEEEauthorblockN{Christopher Frederickson}
\IEEEauthorblockA{Rowan University\\
fredericc0@students.rowan.edu}
\and
\IEEEauthorblockN{Michael Moore}
\IEEEauthorblockA{Rowan University\\
moorem6@students.rowan.edu}
\and
\IEEEauthorblockN{Glenn Dawson}
\IEEEauthorblockA{Rowan University\\
dawson05@students.rowan.edu}
\and
\IEEEauthorblockN{Robi Polikar}
\IEEEauthorblockA{Rowan University\\
polikar@rowan.edu }}

\maketitle

\begin{abstract}
As the prevalence and everyday use of machine learning algorithms, along with our reliance on these algorithms grow dramatically, so do the efforts to attack and undermine these algorithms with malicious intent, resulting in a growing interest in \textit{adversarial machine learning}. A number of approaches have been developed that can render a machine learning algorithm ineffective through poisoning or other types of attacks. Most attack algorithms typically use sophisticated optimization approaches, whose objective function is designed to cause maximum damage with respect to accuracy and performance of the algorithm with respect to some task. In this effort, we show that while such an objective function is indeed brutally effective in causing maximum damage on an embedded feature selection task, it often results in an attack mechanism that can be easily detected with an embarrassingly simple novelty or outlier detection algorithm. We then propose an equally simple yet elegant solution by adding a regularization term to the attacker's objective function that penalizes outlying attack points.

\end{abstract}

\IEEEpeerreviewmaketitle

\section{Introduction} \label{sec:introduction}

Machine learning (ML) algorithms are being applied to an ever-growing spectrum of applications, with dramatic impact of which the general public is largely unaware. Even a simple task of ordering a book online involves  machine learning at multiple stages of the process: from the web search finding the retailer \cite{boyan1996machine}, to advertisements shown alongside the results \cite{miralles2018novel} and recommended products on the website \cite{ziegler2004taxonomy}, fraud detection from the payment provider \cite{maes2002credit}, and to improving the efficiency of the shipping / logistics \cite{chi2007modeling}. Nontrivial matters are also increasingly entrusted to ML algorithms, such as the justice system determining who gets bail \cite{berk2012criminal} and the Department of Defense investigating their use in national security \cite{gadepally2016recommender}. This increased reliance on ML has vastly raised the sensitivity and concerns towards a possible attack due to increased negative impact of a potential vulnerability. The study of the security of learning models at the intersection of ML and cybersecurity is often referred to as \textit{adversarial machine learning} \cite{huang2011adversarial}.

The birth of adversarial machine learning (AML) is often linked to the usage of statistical classifiers to classify spam emails in the early 2000s \cite{biggio2017wild}. Delvi et al. proposed a cost based attack against Bayesian spam filters \cite{dalvi2004spam}. However, Kearns' 1993 work in computational learning theory studying classification in the presence of malicious noise is, to the best of our knowledge, the first work on machine learning in the presence of an adversary \cite{kearns1993learning}. More recently, the susceptibility of deep learning models to adversarial examples \cite{goodfellow2014explaining} has sparked increased interest in the field.

Barreno et al. introduced a taxonomy of adversarial machine learning attacks to classify attacks along three axes: \textit{influence}, \textit{security violation}, and \textit{specificity} \cite{barreno2006can}. \textbf{Influence} describes the mechanism by which the attacker operates: with \textit{causative} attacks (also known as poisoning attacks), the attacker has control of the future training data; in contrast, \textit{exploratory} attacks (evasion attacks) only exploit misclassification. \textbf{Security violation} describes the goal of the attacker: \textit{integrity} attacks attempt to allow malicious data to slip through (i.e., increase the number of false negatives), while \textit{availability} attacks seek to allow non-malicious data to be classified as malicious (i.e., increase the number of false positives); \textit{privacy} attacks attempt to learn information about the classifier or dataset that should not otherwise be available. In cybersecurity, availability attacks are analogous to denial of service attacks. \textbf{Specificity} describes the set of data that is affected: \textit{targeted} attacks focus on a small set of specific data, while \textit{indiscriminate} attacks focus on a large set of nonspecific data.

Attacks across the entire taxonomic spectrum have been applied to a variety of algorithms and applications. For example, poisoning attacks have been shown to be effective against support vector machines \cite{biggio2012poisoning} and modern deep learning algorithms \cite{munoz2017towards}. Evasion attacks have been applied to linear SVM \cite{biggio2013evasion} and a significant body of work has been shown for developing such attacks against deep learning models \cite{papernot2016practical} \cite{goodfellow2015explaining} \cite{papernot2016distillation}. Poisoning attacks have also been developed against clustering algorithms \cite{biggio2013data}, and have been shown to be effective against feature selection algorithms \cite{xiao2015feature}.

In this work, we show that the common poisoning attacks against embedded feature selection can be easily defeated by novelty and outlier detection algorithms. To combat these methods, we modify the attacker's objective function in order to explicitly control the inherent trade-off between the strength of an attack point and the detectability of the attack, and we evaluate the impact of this modification on multiple real-world datasets. Finally, we discuss the importance of this work and the trade-offs in designing secure systems.

\section{Poisoning Attacks}

\subsection{Notation}

Following the notation used by Xiao et al. \cite{xiao2015feature}, we assume that data are generated from a stationary, i.i.d. process $p : \mathcal{X} \mapsto \mathcal{Y}$ (where $\mathcal{X}$ is the set of all possible input features and $\mathcal{Y}$ is the set of all possible output values), from which a set $\mathcal{D} = \{\bm{x}_i, y_i\}_{i=1}^n$ is drawn, where each sample $\mathcal{D}_i$ comprises a $d$-dimensional feature vector $\bm{x}_i = [x_i^1, ..., x_i^d]^\mathrm{T} \in \mathcal{X}$ and a target variable $y_i \in \mathcal{Y}$. 




\subsection{Attacker Knowledge} \label{sec:attacker_knowldge}

There are three levels of knowledge that an adversary may have when developing an attack against a defender: \textit{perfect knowledge}, \textit{limited knowledge}, and \textit{zero knowledge}. \textbf{Perfect knowledge} (also known as white-box) attacks occur when the adversary knows everything about the model (the defender). While this is often infeasible in practice, it is useful to study in order to understand the worst-case scenario. Furthermore, in cybersecurity, it has been demonstrated that security relying on an adversary's lack of knowledge, i.e., security by obscurity, is ineffective \cite{clavier2007secret}. \textbf{Limited knowledge} (or gray-box) attack occurs when the adversary has some level of knowledge of the model. Under this constraint, one approach is to construct a \textit{surrogate} dataset $\hat{\mathcal{D}} = \{\hat{\bm{x}}_i, \hat{y}_i\}_{i=1}^m$, ideally drawn from the same underlying distribution $p$ from which $\mathcal{D}$ was drawn \cite{biggio2013evasion}. This surrogate dataset can be used to train a surrogate classifier that should be similar to the defender. Knowledge of this surrogate classifier can be used when there is missing knowledge of the defender. \textbf{Zero knowledge} (or black-box) attacks occur when the adversary knows nothing about the model prior to developing their attack.

\subsection{Attack Strategy}

Of various objectives that an adversary may have, such as evading detection and taking advantage of the limitations of the learning algorithm, or violating privacy and learning something about the algorithm or data used to train the algorithm, our focus in this work is on poisoning attacks that add malicious data into the training dataset to poison the algorithm.

Biggio et al. define the optimal attack strategy against a learning algorithm as follows: given the knowledge $\bm{\theta}$ that the attacker knows about the learning model (described in Section \ref{sec:attacker_knowldge}), the attacker modifies some data $\mathcal{A} \sim p$ according to the attacker's capabilities $\Phi$ in order to create a modified set of data $\mathcal{A}' \in \Phi(\mathcal{A})$, known as the \textit{attack points} \cite{biggio2013data}. The theoretical effectiveness of the attack is then calculated using some function $\mathcal{W}(\mathcal{A}'; \theta)$. Therefore, the optimal attack strategy is to maximize $\mathcal{W}$ subject to the adversary's capabilities:

\begin{equation} \label{eq:optimal_attack}
\max_{\mathcal{A}'} \ \mathcal{W}(\mathcal{A}'; \bm{\theta})
\end{equation}
\begin{equation*} \label{eq:constraint}
\ \mathrm{s.t.} \ \mathcal{A}' \in \Phi (\mathcal{A})
\end{equation*}

While this generic strategy may generate attack points with the maximal attack strength -- and maximal damage to the classifier -- a carelessly chosen function $\mathcal{W}$ can lead to the na{\"i}ve generation of attack points that are easy to detect as outliers. Such attack points are therefore ineffective against learning systems that implement even the simplest of countermeasures.

\section{Poisoning Attacks Against Embedded Feature Selection Can Be Easily Defeated} \label{sec:defeated}

Xiao et al. proposed an attack strategy to generate a single attack point $\bm{x}_c$ against embedded feature selection algorithms such as LASSO, ridge regression, and the elastic net, trying to force them to choose a poor set of features, with the ultimate goal of inflicting maximum classification loss on a linear classifier $f(\bm{x})=\bm{w}^T\bm{x}+b$ trained with the features selected by the aforementioned feature selection algorithms \cite{xiao2015feature}. The attacker objective is, therefore:

\begin{equation} \label{eq:max_W_xiao}
\max_{\bm{x}_c} \mathcal{W} = \dfrac{1}{m} \sum_{j=1}^{m} \ell (\hat{y}_j, f(\hat{\bm{x}}_j)) + \lambda \Omega(\bm{w})
\end{equation}

\noindent where $m$ is the number of instances, $\ell$ is the loss function (typically, quadratic loss) that the classifier $f$ seeks to minimize, $\lambda$ is the regularization trade-off parameter, $\Omega(\bm{w})$ is the regularization term ($L_1$ for LASSO, $L_2$ for Ridge, and a weighted sum of $L_1$ and $L_2$ for Elastic Net), and $f$ is learned -- by the attacker -- by minimizing

\begin{equation} \label{eq:minL}
\min_{\bm{w},b} \mathcal{L} = \frac{1}{n} \sum_{i=1}^{n} \ell(\hat{y}_i, f(\hat{\bm{x}}_i)) + \lambda \Omega (\bm{w})
\end{equation}

\noindent on $\hat{\mathcal{D}} \, \cup \, \{\bm{x}_c\}$ (if the attacker has \textit{perfect knowledge} of the defender (i.e., not just the model but also the training data), then it can operate directly on the true training data $\mathcal{D}$, instead of the the surrogate data $\hat{\mathcal{D}}$). This strategy is derived from the optimal attack strategy in Equation \ref{eq:optimal_attack}, where $\mathcal{W}$ is the objective function of regularized linear regression. The resulting attack algorithm computes the gradient of the attacker objective (Equation \ref{eq:max_W_xiao}) to yield:

\begin{equation} \label{eq:poisons_efs_gradient}
\dfrac{\partial \mathcal{W}}{\partial \bm{x}_c} = \dfrac{1}{m} \sum_{j=1}^{m} (f(\hat{\bm{x}}_j) - \hat{y}_j) \left( \hat{x}_j^\mathrm{T} \dfrac{\partial \bm{w}}{\partial \bm{x}_c} + \dfrac{\partial b}{\partial \bm{x}_c}\right) + \lambda \bm{r} \dfrac{\partial \bm{w}}{\partial \bm{x}_c}
\end{equation}

\noindent where $\bm{r} = \frac{\partial \Omega}{\partial \bm{w}}$ ($\bm{r} = \textnormal{sub}(\bm{w})$ for LASSO, $\bm{r} = \bm{w}$ for ridge, and $\bm{r} = \rho \, \textnormal{sub}(\bm{w}) + (1-\rho) \, \bm{w}$ for elastic net. $\textnormal{sub}(\bm{w})$ is the sub-gradient, equal to 1 for each positive element of $\bm{w}$, -1 for each negative element, and 0 for elements that are 0, and $\rho$ is the regularization trade-off term weighting the LASSO and ridge regression terms in the elastic net. 

The original gradient ascent algorithm used in \cite{xiao2015feature} employed a line search to set the step size with a relative tolerance stopping criteria, where the algorithm terminated once the difference in $\mathcal{W}$ between two steps fell below some small constant $\epsilon$. For simplicity, we implement the gradient ascent algorithm using a fixed step size $\sigma$, and run for a fixed number of steps $k$. This modified algorithm is shown in Algorithm \ref{alg:poison_efs}. We note that as part of each gradient ascent step, the attack algorithm alters the optimization direction such that the attack point will remain within the feasible domain $\mathcal{B}$ in Step \ref{alg:feasible_domain}, where $\Pi_{\mathcal{B}}(\bm{x})$ is the boundary projection operator that bounds $\bm{x}$ onto the feasible domain $\mathcal{B}$. 

\begin{figure}
\centering
\includegraphics[width=0.5\textwidth]{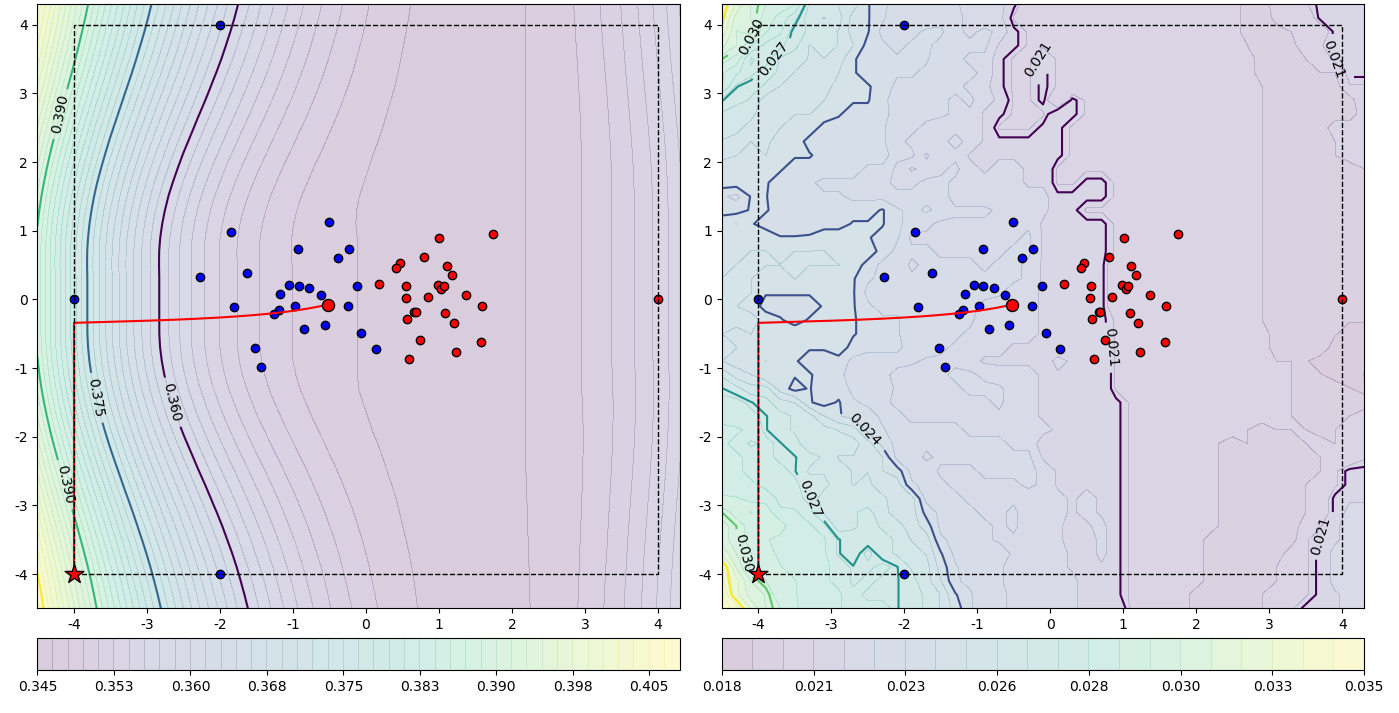}
\caption{Poisoning LASSO regression where $\lambda = 0.01$. This figure is based on our replication of Xiao et al.'s  experiments to recreate Figure 1 in \cite{xiao2015feature}. Red and blue points indicate the two classes. The solid red line indicates the path that the attack point $\bm{x}_c$ took during optimization, following the attacker objective (left), with the star indicating the final attack point. The attacker objective approximates the average classification error over 50 random initializations (right). With each random initialization, 50 instances are sampled from the Gaussian distribution and the attack point pictured appended to poison the dataset. LASSO regression is trained on this poisoned dataset and the classification error is calculated. The border of the feasible domain $\mathcal{B}$ is shown as a dashed line.}
\label{fig:efs_toy_example}
\end{figure}

\begin{algorithm}
  \caption{Poisoning Embedded Feature Selection using Fixed Step Size}
  \label{alg:poison_efs}
  \begin{algorithmic}[1]
    \Require{$\hat{\mathcal{D}}$: surrogate training data}
    \Require{$\{\bm{x}_c^{t=0}, y_c\}_{c=1}^{q}$: $q$ initial attack points with labels}
    \Require{$\sigma$: step size}
    \Require{$k$: number of steps}
    
    \For{$t = 1,\ldots,k$}
      \For{$c = 1,\ldots,q$}
          \State $\{\bm{w},b\} \leftarrow$ learn classifier on $\hat{\mathcal{D}} \cup \{\bm{x}_c^{t-1}\}_{c=1}^q$
          \State Calculate $\nabla \mathcal{W}$ according to Equation \ref{eq:poisons_efs_gradient}
          \State $\bm{d} = \Pi_{\mathcal{B}} \left(\bm{x}_c^{t-1} + \nabla \mathcal{W} \right) - \bm{x}_c^{t-1}$ \label{alg:feasible_domain}
          \State $\bm{x}_c^t = \bm{x}_c^{t-1} + \sigma \bm{d}$
      \EndFor
    \EndFor
    \State
    \Return $\{\bm{x}_c^{t=k}\}_{c=1}^q$ 
  \end{algorithmic}
\end{algorithm}

Given a surrogate dataset $\hat{\mathcal{D}}$ (equivalent to $\mathcal{D}$ in case of perfect knowledge), $q$ attack points are randomly initialized. For each time step $t$ and each attack point $c$, the algorithm (i.e., the attacker) first learns $f$ by minimizing Equation \ref{eq:minL}, and uses it to compute the attacker objective gradient as in Equation \ref{eq:poisons_efs_gradient}. Then, the gradient direction $\bm{d}$ is calculated by projecting the sum of the attack point and the objective gradient into the feasible domain $\mathcal{B}$ before subtracting the original attack point vector. Finally, the updated attack point is calculated by summing the original attack point vector with the product of the gradient direction and the step size $\sigma$.

An example of this attack algorithm applied to a two-dimensional Gaussian toy dataset is shown in Figure \ref{fig:efs_toy_example}, for which we implemented and replicated Xiao et al.'s experiment in \cite{xiao2015feature}. A careful observation of Figure \ref{fig:efs_toy_example} shows that the final attack point lies on the border of the feasible space bounded by $\Pi_{\mathcal{B}}$ -- an arbitrary boundary set by the attacker with no justification with respect to underlying application.

For some applications, the constraints on input features may be well defined. For example, a pixel in an image has a clear boundary, and a limited range of color values. However, many other problems have no clearly-set boundaries. For example, a bag-of-words model for spam filtering, where each feature is the count of each word, has no effective upper bound.

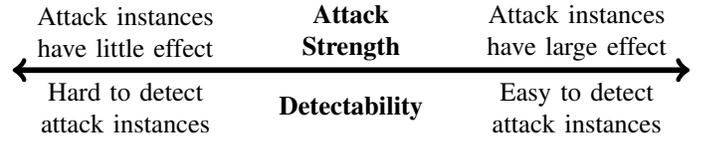
\begin{figure}
\centering
\begin{tikzpicture}

\node[align=center] at (0,0.5) {\textbf{Attack} \\ \textbf{Strength}};
\node[align=center] at (-3,0.5) {Attack instances \\ have little effect};
\node[align=center] at (3,0.5) {Attack instances \\ have large effect};

\draw[<->, line width=0.75mm] (-4.5,0) -- (4.5,0);

\node[align=center] at (0,-0.5) {\textbf{Detectability}};
\node[align=center] at (-3,-0.5) {Hard to detect \\ attack instances};
\node[align=center] at (3,-0.5) {Easy to detect \\ attack instances};

\end{tikzpicture}
\caption{The attack strength-detectability continuum. A trade-off must be made between the impact of the attack instances and how easy they are to detect.}
\label{fig:attack_strength_detectibility}
\end{figure}

Additionally, because the attack points fall along the boundary of the feasible space, the boundary itself plays a critical role in both the effectiveness and detectability of the attack points. If the boundary is set such that it remains well within the convex hull of the dataset, the generated attack points can blend in with legitimate data, and not be easily detectable as malicious, but then the attack will have little impact on the defender. If a wider boundary is used, well outside the convex hull of the data, the attack can have a large negative impact on the defender, but the data may be easily detectable as malicious with a simple outlier detection. This attack strength vs. detectability dilemma is illustrated in Figure \ref{fig:attack_strength_detectibility}.

\section{Updating The Attack Objective with an Outlier Detection Evasion Term} \label{sec:updated}


We argue that the effectiveness of a poisoning attack drops to zero if the poisoned data is easily detected and removed, and that it is necessary to consider the detectability of an attack point as well as the theoretical optimality of the poisoned data. Therefore, in order to generate attack points that simultaneously maximize the impact  the learner's objective function and minimize the defender's capability for detection, we modify the attack strategy in Equation \ref{eq:max_W_xiao} by adding a penalty term:

\begin{equation}
\max_{\bm{x}_c} \mathcal{W}' = \mathcal{W} - \phi \Lambda (\bm{x}_c)
\end{equation}

\noindent where $\Lambda$ is a function of \textit{outlier detectability} and $\phi$ is a weighting term. The addition of this term modifies the gradient of the attack strategy from Equation \ref{eq:poisons_efs_gradient} to:

\begin{equation}
\dfrac{\partial \mathcal{W}'}{\partial \bm{x}_c} = \dfrac{\partial \mathcal{W}}{\partial \bm{x}_c} - \phi \dfrac{\partial \Lambda}{\partial \bm{x}_c}
\end{equation}

\noindent Adding the $\Lambda$ term to the attacker objective function allows attack points to be generated anywhere along the attack strength-versus-detectability \textit{continuum}, shown in Figure \ref{fig:attack_strength_detectibility}, depending on the outlier detection countermeasures used by the defender. 


Ideally, $\Lambda$ should be selected to directly oppose the outlier detection algorithm used by the defender; in practice, it is very difficult or impossible to know precisely what defensive methods a defender may be using. Therefore, it is necessary to select a surrogate outlier detection algorithm against which to optimize $\Lambda$. Here, we show the calculation of $\Lambda$ against two possible surrogate outlier detection algorithms: distance threshold and $k$-th nearest neighbor, chosen based on the ease of gradient computation. If the defender is known to be using a more sophisticated outlier detection algorithm, a more complex gradient calculation may be needed. 

\subsection{Distance Threshold}

\begin{figure}
\centering
\includegraphics[width=0.38\textwidth]{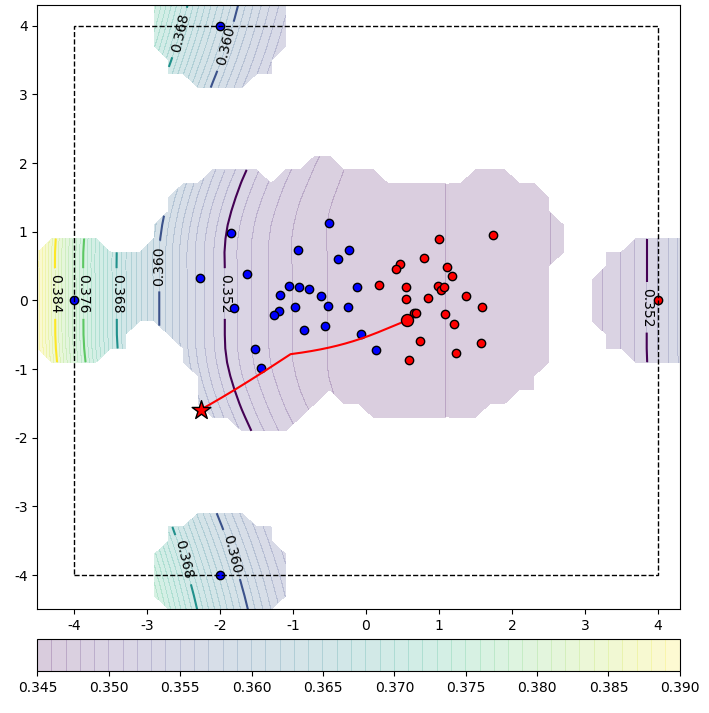}
\caption{The attacker objective function of the modified attack using attacker distance threshold of $d_{att}$=1 on a 2D toy Gaussian dataset. The red star is the final attack instance. When the distance from the attack instance to any other instance is less than 1, the attacker objective function is identical to that of the original attack. When the distance from the attack instance to any other instance is greater than 1, the optimization stops. The white region corresponds to $D(\bm{x}_c, \bm{x}_k) > d_{att} \rightarrow \Lambda (\bm{x}_c) = \infty$, and essentially determines the data-driven boundary of the feasible space}.
\label{fig:lasso_distance_threshold}
\end{figure}

The distance threshold method defines an outlier with respect to the distance $D$ between the attack point $\bm{x}_c$ and its nearest instance $\bm{x}'$ in the dataset. The attacker, knowing or suspecting that the defender may be using outlier detection, chooses an attack threshold $d_{att}$ that is smaller than the defender threshold $d_{def}$ it thinks the defender is using. Then, if the distance $D$ is above the threshold $d_{att}$, the attacker knows that the attack point will be detected as an outlier. Hence, from the attacker's perspective, the $\Lambda$ term can be defined as

\begin{equation}
  \Lambda(\bm{x}_c) =
  \begin{cases} 
      \infty & D(\bm{x}_c, \bm{x}') > d_{att} \\
      0 & otherwise
   \end{cases}
\end{equation}

By this definition, the gradient calculation is identical to Equation \ref{eq:poisons_efs_gradient} when the distance $D < d_{att}$. However, when the attack point moves outside of the attacker distance threshold, the penalty is set to infinity, preventing the algorithm from continuing outside of the distance boundary. With this approach, the attack strength is controlled by varying the attacker distance threshold parameter; hence $d_{att}$ effectively serves as the outlier weight term $\phi$. An example of an attack point generated using this outlier term is shown in Figure \ref{fig:lasso_distance_threshold}. Note that the attack point is now on a data-driven boundary (the edge of white regions), as opposed to user-defined one (outer boundary shown as the dashed line as was the case in \cite{xiao2015feature}).  

\subsection{k-th Nearest Neighbor}

\begin{figure}
\centering
\includegraphics[width=0.38\textwidth]{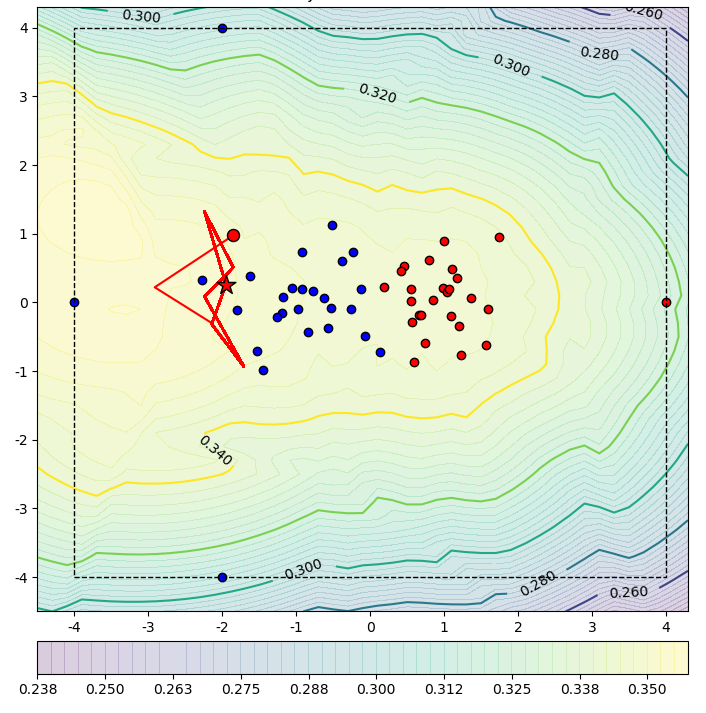}
\caption{The attacker objective function of the modified attack using the k-nearest neighbor outliers term with $\phi = 0.005, P = 2, k = 3$ on a 2D toy Gaussian dataset. Due to the discontinuity of the objective function, the attack instance becomes stuck in a local maximum.}
\label{fig:lasso_k_nearest_neighbor}
\end{figure}

Another choice of outlier term is related to the distance to the $k$-th nearest neighbor $\bm{x}_k$ raised to some power.

\begin{equation}
  \Lambda(\bm{x}_c) = \left\lVert \bm{x}_c - \bm{x}_k \right\rVert_2^P
\end{equation}

\noindent where $\bm{x}_c$ is the attack point, $\bm{x}_k$ is the $k$-th nearest point $\in \mathcal{D}$, and $P$ is a user-defined parameter. The resulting gradient is 

\begin{equation}
  \dfrac{\partial \Lambda}{\partial \bm{x}_c} = P \cdot \left\lVert \bm{x}_c - \bm{x}_k \right\rVert_2^{P-2} \cdot \left(\bm{x}_c - \bm{x}_k\right)
\end{equation}

An example of an attack point generated using this outlier term is shown in Figure \ref{fig:lasso_k_nearest_neighbor}. However, this term has a significant drawback: the objective function becomes discontinuous due to the abruptly changing $k$-th nearest neighbor, resulting in the attack point often getting stuck in a local maximum. Because of this tendency, we do not examine the case of assuming this outlier detection method in our experiments.



\section{Experimental Methods and Results} \label{sec:methods_results}

\subsection{Datasets} \label{sec:datasets}

To evaluate the effectiveness of both the original attacks, described in \cite{xiao2015feature}, and the proposed attacks augmented with $\Lambda$, we attack LASSO regression trained on three UCI datasets: \textit{spambase}, \textit{credit approval}, and \textit{congressional voting} \cite{Lichman:2013}. 

The \textbf{spambase} dataset contains both \textit{spam} and \textit{non-spam} (i.e., \textit{ham}) emails, where 48 features are the percentage of words in the email that are a particular word (e.g., address, free, business, etc.), six features are the percentage of characters in the email that are a particular character (e.g., \#, \$, !), one feature is the average length of uninterrupted capital letters, one feature is the maximum length of uninterrupted capital letters, and one feature is the sum of all uninterrupted capital letter sequences. The feasible boundary of the first 54 features is $[0, 100]$; and is $[0, \infty]$ for the last three. 

The \textbf{credit approval} dataset contains features about individuals that either were or were not granted credit from a Japanese credit approval company. It contains 15 features: 6 are continuous, whose values range from 0 to 100,000; 9 are categorical, ranging from 2 to 14 categories. The exact meaning of each feature is not stated and therefore the true boundaries of the continuous input features are unknown; however, at least one feature is related to the individual's income, which has no set upper bound. 

The \textbf{congressional voting} dataset contains the voting records of 435 members of the U.S. House of Representative in 1984. The 16 features are binary values that represent whether they voted for or against a bill on certain topics (e.g., El Salvador aid, religious groups in school, anti-satellite weapon test ban, etc.). The objective is to predict the congressperson's political affiliation, either Democrat or Republican.

All features are normalized such that their mean is zero and standard deviation is 1. 80\% of the data is used as the training dataset, while 20\% reserved for testing.

\subsection{Novelty and Outlier Detection Methods}

For these experiments, the LASSO regression classifiers trained on the UCI datasets were augmented with four novelty and outlier detection algorithms: \textit{distance threshold}, \textit{one-class SVM}, \textit{isolation forest}, and \textit{local outlier factor}, with each algorithm being applied separately. We used our own implementation of the distance threshold algorithm, while the implementations of the one-class SVM, isolation forest, and local outlier factor algorithms were obtained from the {\tt\footnotesize scikit-learn} library \cite{scikit-learn}.

\subsubsection{Distance Threshold} \label{sec:nad_distance_thresold}

Distance threshold is the simplest method of outlier detection: if the Euclidean distance between a new instance $\bm{x}_c$ and its nearest neighbor $\bm{x}_i \in \hat{\mathcal{D}}$ is less than some defender distance threshold $d_{def}$, then the new instance is added to the dataset on which the classifier is trained; otherwise, the new instance is considered an outlier and discarded, and the dataset remains unchanged. It is important to note that while they serve similar purposes, the defender distance threshold $d_{def}$ is different from the attacker distance threshold $d_{att}$: the defender distance threshold determines the defender's sensitivity to \textit{identifying} data as outliers; the attacker distance threshold determines the extent to which the attack instances can be hidden from being detected as outliers.


\subsubsection{One-Class SVM} \label{sec:nad_one_class_svm}

One-class support vector machines act as one-versus-all classifiers, drawing a classification boundary between members of a specific class (legitimate data) and all other data (i.e., poisoned attack data). To accomplish this, the optimization function differs from the traditional two-class SVM in that there is no opposing class data between which to draw an optimal hyperplane; instead, the goal is to maximize the distance between the classification hyperplane and the origin in the appropriate high-dimensional feature space \cite{scholkopf2000support}. Any data point that lies inside of this hyperplane is considered as a legitimate data point, and any data point that lies outside of this hyperplane is considered an outlier.

\subsubsection{Isolation Forest} \label{sec:nad_isolation_forest}


Isolation forests seek to exploit the fact that outlier data will generally exhibit two distinctive properties: i.) they are the minority data, and ii.) they will have features that are distinct from the clustered data \cite{liu2008isolation}. By constructing a decision tree focused on isolating these outlier points rather than categorizing normal data points, the Isolation Tree can distinguish outliers by measuring the depth of the leaf node containing each data point: outliers will be categorized and isolated much closer to the root node than the more clustered normal data. An Isolation Forest is simply an ensemble of Isolation Trees, which achieves a higher performance than a single tree alone.

\subsubsection{Local Outlier Factor} \label{sec:nad_local_outlier_factor} 


Local outlier factor is defined as the ratio between an instance's density compared to the density of the $k$-nearest neighbors. If an instance is an outlier, then the local density is expected to be much lower than the density of the $k$-nearest neighbors \cite{breunig2000lof}.

\subsection{Experiment 1: Original Attack Performance Against Novelty and Outlier Detection}

\textit{The Experiment:} For each training dataset, two attack instances are generated using the original attack method described in \cite{xiao2015feature} using the same $\lambda = 0.1$. These two attack instances are combined with 40 instances from the test dataset. Outlier scores of the poisoned datasets are calculated using the four novelty and outlier detection methods described in Section \ref{sec:defeated}. 

\textit{Results:}
Figure \ref{fig:efs_outlier} shows the sorted outlier scores of the poisoned dataset; the attack instances are shown in red, while the legitimate instances from the test dataset are shown in blue. In all experiments, at least one novelty or outlier detection algorithm gave the attack instances the highest outlier scores, with the isolation forest, distance threshold, and one-class support vector machine giving the attack instances the highest outlier score in all cases. This simple experiment shows that, in a practical scenario, the original attack algorithm described in \cite{xiao2015feature} will generate attack instances that are easily identified as outliers.

\subsection{Experiment 2: Improved Attack Incorporating Novelty and Outlier Detection Evasion}

\textit{The Experiment:}
As in Experiment 1, two attack instances are generated for each training dataset using the modified attack method described in Section \ref{sec:updated} with the same $\lambda = 0.1$ however now with the attacker distance threshold parameter $d_{att}$ set to 1. These two attack instances are combined with 40 instances from the test dataset. The outlier scores of the poisoned datasets are calculated using the four novelty and outlier detection methods described in Section \ref{sec:defeated}.

\textit{Results:}
Figure \ref{fig:efs_outlier_distance_threshold} shows the sorted outlier scores of the poisoned dataset. In just about all cases, the attack points were not given the highest outlier score. In fact, in most cases the attack points received much smaller outlier scores, indicating that they likely would not have been detected and removed as outliers. This experiment demonstrates that it is possible to generate attack instances that can evade outlier detection against classifiers trained on real-world datasets.


\subsection{Experiment 3: Distance Threshold Based Countermeasures on Toy Gaussian Dataset}

\textit{The Experiment:}
While we have shown that the attack instances from the original attack are easily identified as outliers, and that the instances from the modified attack can be created such that they are not identified as outliers, there is a wide spectrum of attacks that can be chosen, with varying attack strengths. From the attacker's perspective, the goal is to select the attack with the highest attack strength that is not detected by any of the defender's countermeasures; from the defender's perspective, the goal is to select a defense that will block attack instances while still allowing legitimate data.

Using LASSO regression with $\lambda = 0.1$, we train three classifiers on the toy Gaussian dataset from Figure \ref{fig:efs_toy_example}, each being augmented with a distance threshold outlier detector with $d_{def}$ set to 1, 3, or 5. As a control, we also train the classifier using LASSO regression using no outlier detection. Attack instances are generated using our improved attack algorithm, with \textit{attacker} distance thresholds varying between 0 and 10. At each such threshold, a poisoned dataset of 50 non-malicious instances and 1 attack instance is generated. This is repeated 50 times and the average classification accuracy on the poisoned data is calculated for each defender.

\pgfplotsset{compat = 1.3}
\begin{figure}[t]
\begin{tikzpicture}
    \begin{groupplot}[
    	group style = {group size = 4 by 3,
        				vertical sep=15pt,
        				horizontal sep=15pt
                        }, 
        width = 0.35\linewidth, 
        height = 2.7cm,
        axis line style=black,
        yticklabel style = {font=\tiny,xshift=0.5ex},
        ylabel style = {font=\tiny,yshift=-0.1cm},
        xticklabel style = {font=\tiny,yshift=0.5ex}
        ]
        
\nextgroupplot[title=DT, ylabel=spambase]
\addplot[ycomb,color=spprl-blue] table[x index = {0}, y index = {1},col sep=comma] {data-f1/f1_norm_dt_spambase.csv};
\addplot[ycomb,color=spprl-red] table[x index = {0}, y index = {1},col sep=comma] {data-f1/f1_out_dt_spambase.csv};

\nextgroupplot[title=1C-SVM]
\addplot[ycomb,color=spprl-blue] table[x index = {0}, y index = {1},col sep=comma] {data-f1/f1_norm_svm_spambase.csv};
\addplot[ycomb,color=spprl-red] table[x index = {0}, y index = {1},col sep=comma] {data-f1/f1_out_svm_spambase.csv};

\nextgroupplot[title=IF]
\addplot[ycomb,color=spprl-blue] table[x index = {0}, y index = {1},col sep=comma] {data-f1/f1_norm_if_spambase.csv};
\addplot[ycomb,color=spprl-red] table[x index = {0}, y index = {1},col sep=comma] {data-f1/f1_out_if_spambase.csv};

\nextgroupplot[title=LOF]
\addplot[ycomb,color=spprl-blue] table[x index = {0}, y index = {1},col sep=comma] {data-f1/f1_norm_lof_spambase.csv};
\addplot[ycomb,color=spprl-red] table[x index = {0}, y index = {1},col sep=comma] {data-f1/f1_out_lof_spambase.csv};

\nextgroupplot[ylabel=credit-approval]
\addplot[ycomb,color=spprl-blue] table[x index = {0}, y index = {1},col sep=comma] {data-f1/f1_norm_dt_credit_approval.csv};
\addplot[ycomb,color=spprl-red] table[x index = {0}, y index = {1},col sep=comma] {data-f1/f1_out_dt_credit_approval.csv};

\nextgroupplot
\addplot[ycomb,color=spprl-blue] table[x index = {0}, y index = {1},col sep=comma] {data-f1/f1_norm_svm_credit_approval.csv};
\addplot[ycomb,color=spprl-red] table[x index = {0}, y index = {1},col sep=comma] {data-f1/f1_out_svm_credit_approval.csv};

\nextgroupplot
\addplot[ycomb,color=spprl-blue] table[x index = {0}, y index = {1},col sep=comma] {data-f1/f1_norm_if_credit_approval.csv};
\addplot[ycomb,color=spprl-red] table[x index = {0}, y index = {1},col sep=comma] {data-f1/f1_out_if_credit_approval.csv};

\nextgroupplot
\addplot[ycomb,color=spprl-blue] table[x index = {0}, y index = {1},col sep=comma] {data-f1/f1_norm_lof_credit_approval.csv};
\addplot[ycomb,color=spprl-red] table[x index = {0}, y index = {1},col sep=comma] {data-f1/f1_out_lof_credit_approval.csv};

\nextgroupplot[ylabel=congressional-voting]
\addplot[ycomb,color=spprl-blue] table[x index = {0}, y index = {1},col sep=comma] {data-f1/f1_norm_dt_congressional_voting.csv};
\addplot[ycomb,color=spprl-red] table[x index = {0}, y index = {1},col sep=comma] {data-f1/f1_out_dt_congressional_voting.csv};

\nextgroupplot
\addplot[ycomb,color=spprl-blue] table[x index = {0}, y index = {1},col sep=comma] {data-f1/f1_norm_svm_congressional_voting.csv};
\addplot[ycomb,color=spprl-red] table[x index = {0}, y index = {1},col sep=comma] {data-f1/f1_out_svm_congressional_voting.csv};

\nextgroupplot
\addplot[ycomb,color=spprl-blue] table[x index = {0}, y index = {1},col sep=comma] {data-f1/f1_norm_if_congressional_voting.csv};
\addplot[ycomb,color=spprl-red] table[x index = {0}, y index = {1},col sep=comma] {data-f1/f1_out_if_congressional_voting.csv};

\nextgroupplot
\addplot[ycomb,color=spprl-blue] table[x index = {0}, y index = {1},col sep=comma] {data-f1/f1_norm_lof_congressional_voting.csv};
\addplot[ycomb,color=spprl-red] table[x index = {0}, y index = {1},col sep=comma] {data-f1/f1_out_lof_congressional_voting.csv};

	\end{groupplot}
\end{tikzpicture}
\caption{The sorted outlier scores of the attack instances generated by the original attack algorithm in \cite{xiao2015feature} (red) and  40 legitimate instances (blue) from the spambase, congressional voting, and credit approval datasets. The outlier scores are computed using the distance threshold, one-class SVM, local outlier factor, and isolation forest novelty and outlier detection algorithms. For all datasets, at least one of the novelty or outlier detection algorithms gave the attack instances the highest outlier score.}
\label{fig:efs_outlier}
\end{figure}
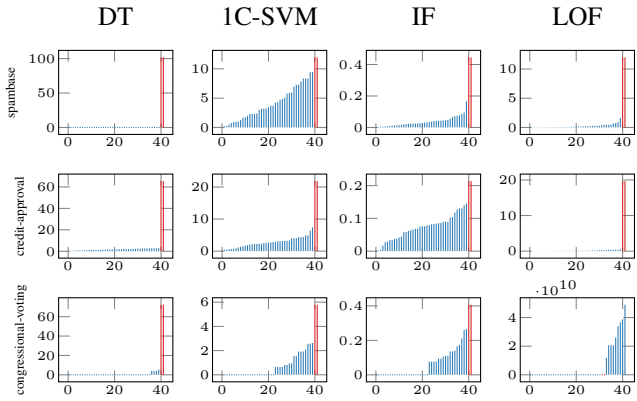

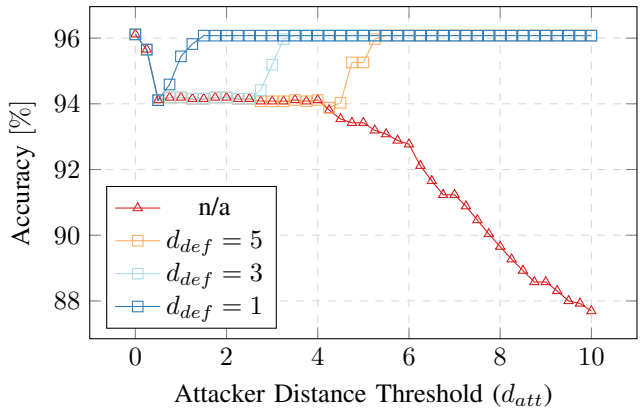
\begin{figure}
\vspace{0.1in}
\centering
\begin{tikzpicture}
\begin{axis}[
		width = 1\linewidth, 
        height = 6.0cm,
        grid=major,
        grid style={dashed,gray!30},
        xlabel=Attacker Distance Threshold ($d_{att}$),
        ylabel=Accuracy,
        y unit=\si{\percent},
        axis line style=black,
        legend style={at={(0.03,0.03)},anchor=south west},
        ]
\addplot[color=spprl-red,mark=triangle] 
        table[x=distance_threshold,y=model_1,col sep=comma] {data/attack_strength.csv}; \addlegendentry{n/a}
\addplot[color=spprl-light-orange,mark=square] 
        table[x=distance_threshold,y=model_2,col sep=comma] {data/attack_strength.csv}; \addlegendentry{$d_{def} = 5$}
\addplot[color=spprl-light-blue,mark=square] 
        table[x=distance_threshold,y=model_3,col sep=comma] {data/attack_strength.csv}; \addlegendentry{$d_{def} = 3$}
\addplot[color=spprl-blue,mark=square] 
        table[x=distance_threshold,y=model_4,col sep=comma] {data/attack_strength.csv}; \addlegendentry{$d_{def} = 1$}
\end{axis}
\end{tikzpicture}
\caption{The impact of applying a distance threshold based countermeasure with varying defender distance thresholds ($d_{def} = 1,3,5$) to a LASSO regression classifier ($\lambda = 0.1$) trained on a two-class Gaussian dataset. The dataset is poisoned by adding a single attack instance to the dataset generated using an attacker distance threshold between 0 and 10. The average accuracy over 50 random runs is reported.}
\label{fig:attack_strength_curve}
\end{figure}

\textit{Results:}
Figure \ref{fig:attack_strength_curve} shows the average classification error vs. attacker distance threshold on all four cases. With no countermeasure ($d_{def}=0$), the single attack point causes a significant drop in accuracy from 96.1\% to 87.8\%; a remarkable drop considering that the single attack point constitutes less than 2\% of the augmented dataset. Adding outlier detection reduces the negative impact of the attack instance: when the defender distance threshold is 1, 3, and 5, the accuracy drops only to 94.1\%, 94.2\%, and 93.9\%, respectively. As the attacker's outlier term is identical to the outlier detection mechanism used by the defender (because the attacker has full or partial knowledge of the defender), the distance thresholds chosen by the attacker and defender are related. The relationship between the two distance thresholds, and its effect on the classifier's performance, is worth noting: when the attacker threshold is less than that of the defender, the attack instances are not detected as outliers and have the maximum impact; when the attacker threshold is greater than that of the defender, the attack instances are detected and removed from the dataset, and the attack is completely eliminated.

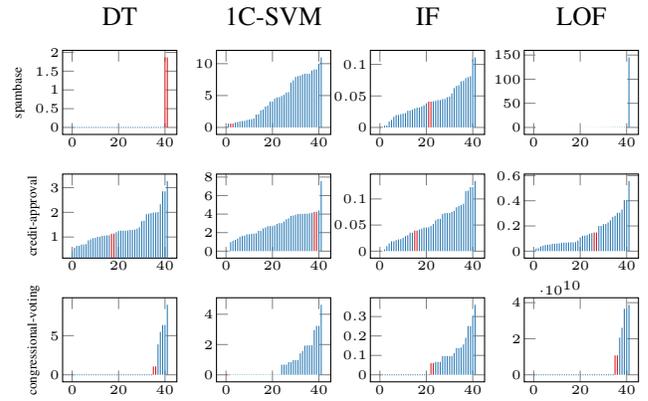
\begin{figure}
\begin{tikzpicture}
    \begin{groupplot}[
    	group style = {group size = 4 by 3,
        				vertical sep=15pt,
        				horizontal sep=15pt
                        }, 
        width = 0.35\linewidth, 
        height = 2.7cm,
        axis line style=black,
        yticklabel style = {font=\tiny,xshift=0.5ex, /pgf/number format/fixed},
        ylabel style = {font=\tiny,yshift=-0.1cm},
        xticklabel style = {font=\tiny,yshift=0.5ex}
        ]
        
\nextgroupplot[title=DT,ylabel=spambase]
\addplot[ycomb,color=spprl-blue] table[x index = {0}, y index = {1},col sep=comma] {data-f2/f2_norm_dt_spambase.csv};
\addplot[ycomb,color=spprl-red] table[x index = {0}, y index = {1},col sep=comma] {data-f2/f2_out_dt_spambase.csv};

\nextgroupplot[title=1C-SVM]
\addplot[ycomb,color=spprl-blue] table[x index = {0}, y index = {1},col sep=comma] {data-f2/f2_norm_svm_spambase.csv};
\addplot[ycomb,color=spprl-red] table[x index = {0}, y index = {1},col sep=comma] {data-f2/f2_out_svm_spambase.csv};

\nextgroupplot[title=IF]
\addplot[ycomb,color=spprl-blue] table[x index = {0}, y index = {1},col sep=comma] {data-f2/f2_norm_if_spambase.csv};
\addplot[ycomb,color=spprl-red] table[x index = {0}, y index = {1},col sep=comma] {data-f2/f2_out_if_spambase.csv};

\nextgroupplot[title=LOF]
\addplot[ycomb,color=spprl-blue] table[x index = {0}, y index = {1},col sep=comma] {data-f2/f2_norm_lof_spambase.csv};
\addplot[ycomb,color=spprl-red] table[x index = {0}, y index = {1},col sep=comma] {data-f2/f2_out_lof_spambase.csv};

\nextgroupplot[ylabel=credit-approval]
\addplot[ycomb,color=spprl-blue] table[x index = {0}, y index = {1},col sep=comma] {data-f2/f2_norm_dt_credit_approval.csv};
\addplot[ycomb,color=spprl-red] table[x index = {0}, y index = {1},col sep=comma] {data-f2/f2_out_dt_credit_approval.csv};

\nextgroupplot
\addplot[ycomb,color=spprl-blue] table[x index = {0}, y index = {1},col sep=comma] {data-f2/f2_norm_svm_credit_approval.csv};
\addplot[ycomb,color=spprl-red] table[x index = {0}, y index = {1},col sep=comma] {data-f2/f2_out_svm_credit_approval.csv};

\nextgroupplot
\addplot[ycomb,color=spprl-blue] table[x index = {0}, y index = {1},col sep=comma] {data-f2/f2_norm_if_credit_approval.csv};
\addplot[ycomb,color=spprl-red] table[x index = {0}, y index = {1},col sep=comma] {data-f2/f2_out_if_credit_approval.csv};

\nextgroupplot
\addplot[ycomb,color=spprl-blue] table[x index = {0}, y index = {1},col sep=comma] {data-f2/f2_norm_lof_credit_approval.csv};
\addplot[ycomb,color=spprl-red] table[x index = {0}, y index = {1},col sep=comma] {data-f2/f2_out_lof_credit_approval.csv};

\nextgroupplot[ylabel=congressional-voting]
\addplot[ycomb,color=spprl-blue] table[x index = {0}, y index = {1},col sep=comma] {data-f2/f2_norm_dt_congressional_voting.csv};
\addplot[ycomb,color=spprl-red] table[x index = {0}, y index = {1},col sep=comma] {data-f2/f2_out_dt_congressional_voting.csv};

\nextgroupplot
\addplot[ycomb,color=spprl-blue] table[x index = {0}, y index = {1},col sep=comma] {data-f2/f2_norm_svm_congressional_voting.csv};
\addplot[ycomb,color=spprl-red] table[x index = {0}, y index = {1},col sep=comma] {data-f2/f2_out_svm_congressional_voting.csv};

\nextgroupplot
\addplot[ycomb,color=spprl-blue] table[x index = {0}, y index = {1},col sep=comma] {data-f2/f2_norm_if_congressional_voting.csv};
\addplot[ycomb,color=spprl-red] table[x index = {0}, y index = {1},col sep=comma] {data-f2/f2_out_if_congressional_voting.csv};

\nextgroupplot
\addplot[ycomb,color=spprl-blue] table[x index = {0}, y index = {1},col sep=comma] {data-f2/f2_norm_lof_congressional_voting.csv};
\addplot[ycomb,color=spprl-red] table[x index = {0}, y index = {1},col sep=comma] {data-f2/f2_out_lof_congressional_voting.csv};

	\end{groupplot}
\end{tikzpicture}
\caption{The sorted outlier scores of the attack instances computed from the modified attack algorithm, using a distance threshold outlier term with an attacker distance threshold $d_{att}$ of 1 (red), injected into a set of 40 legitimate instances (blue) from the spambase, congressional voting, and credit approval datasets. The outlier scores are computed using the distance threshold, one-class support vector machine, local outlier factor, and isolation forest. In most cases, the attack instances were not given the highest outlier score and likely would not have been labeled as outliers.}
\label{fig:efs_outlier_distance_threshold}
\end{figure}

Using this modified attack, the attacker can select an appropriate threshold with respect to the countermeasures used by the defender, based on whether the attacker has full or partial knowledge of the defender (model). In this particular case, it is advantageous for the defender to use an outlier detection algorithm that is very sensitive to outliers (i.e., a low $d_{def}$). However, the improved outlier detection comes at the cost of making the classifier less able to deal with drift or covariate shift as any drift will also be detected as outliers. 

\subsection{Experiment 4: Impact of Attacker Distance Threshold on Detectability}

\textit{The Experiment:}
Using the datasets described in Section \ref{sec:datasets}, we evaluate the impact of $d_{att}$ on detectability. For each dataset-outlier detector pair, we generate attack instances using the proposed attack mechanism with attacker distance thresholds $d_{att}$ varying between 0 and 10. At each attacker distance threshold, we calculate the outlier score, evaluated on a poisoned dataset containing the  legitimate test data augmented with the appropriate attack instance.

\begin{figure}[t]
\begin{tikzpicture}
    \begin{groupplot}[
    	group style = {group size = 4 by 3,
        				vertical sep=15pt,
        				horizontal sep =15pt
                        }, 
        width = 0.35\linewidth, 
        height = 2.7cm,
        grid=major,
        grid style={dashed,gray!30},
        axis line style=black,
        yticklabel style = {font=\tiny,xshift=0.5ex, /pgf/number format/fixed},
        ylabel style = {font=\tiny,yshift=-0.1cm},
        xticklabel style = {font=\tiny,yshift=0.5ex},
        xlabel style = {font=\tiny,yshift=0.1cm},
        ]
        
\nextgroupplot[title=DT,ylabel=spambase]
\addplot[color=spprl-blue] table[x index = {0}, y index = {1},col sep=space] {data-f3/f3_dt_spambase.csv};

\nextgroupplot[title=1C-SVM]
\addplot[color=spprl-blue] table[x index = {0}, y index = {1},col sep=space] {data-f3/f3_svm_spambase.csv};

\nextgroupplot[title=IF]
\addplot[color=spprl-blue] table[x index = {0}, y index = {1},col sep=space] {data-f3/f3_if_spambase.csv};

\nextgroupplot[title=LOF]
\addplot[color=spprl-blue] table[x index = {0}, y index = {1},col sep=space] {data-f3/f3_lof_spambase.csv};

\nextgroupplot[ylabel=credit-approval]
\addplot[color=spprl-blue] table[x index = {0}, y index = {1},col sep=space] {data-f3/f3_dt_credit_approval.csv};

\nextgroupplot
\addplot[color=spprl-blue] table[x index = {0}, y index = {1},col sep=space] {data-f3/f3_svm_credit_approval.csv};

\nextgroupplot
\addplot[color=spprl-blue] table[x index = {0}, y index = {1},col sep=space] {data-f3/f3_if_credit_approval.csv};

\nextgroupplot
\addplot[color=spprl-blue] table[x index = {0}, y index = {1},col sep=space] {data-f3/f3_lof_credit_approval.csv};

\nextgroupplot[ylabel=congressional-voting,xlabel=Attacker Distance Threshold $d_{att}$]
\addplot[color=spprl-blue] table[x index = {0}, y index = {1},col sep=space] {data-f3/f3_dt_congressional_voting.csv};

\nextgroupplot
\addplot[color=spprl-blue] table[x index = {0}, y index = {1},col sep=space] {data-f3/f3_svm_congressional_voting.csv};

\nextgroupplot
\addplot[color=spprl-blue] table[x index = {0}, y index = {1},col sep=space] {data-f3/f3_if_congressional_voting.csv};

\nextgroupplot
\addplot[color=spprl-blue] table[x index = {0}, y index = {1},col sep=space] {data-f3/f3_lof_congressional_voting.csv};

	\end{groupplot}
\end{tikzpicture}
\caption{Outlier score vs attacker distance threshold on the spambase, credit approval, and congressional voting datasets using distance threshold (DT), one-class support vector machines (1C-SVM), isolation forests (IF), and local outlier factor (LOF). In general, as the attacker distance threshold increases, the outlier score also increases and the attack instances are more easily detected.}
\label{fig:outlier_vs_attack_strength}
\end{figure}
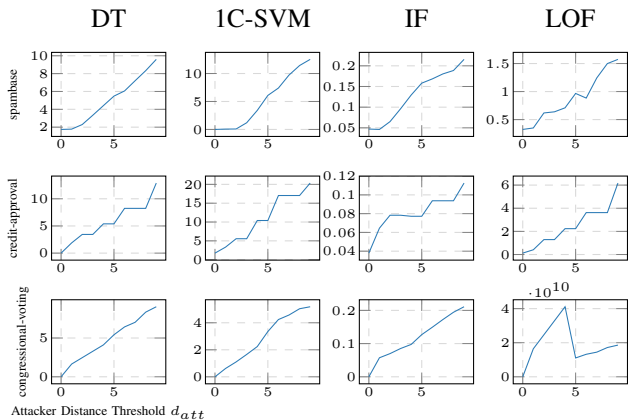

\textit{Results:}
Figure \ref{fig:outlier_vs_attack_strength} shows that for every dataset-outlier detector pair there is a clear trade-off between outlier score and attack strength. Although a higher outlier score corresponds to a greater impact on the defender's loss, it also results in a higher detectability, which could lead to zero impact if the attack point is detected and removed.

\section{Discussion} \label{sec:discussion}

Using real datasets, we have shown that a clear trade-off exists between the impact (i.e., damage) of an attack instance and its detectability by novelty and outlier detection techniques. While our experiments were on LASSO, there is no reason to believe that this trend between impact and detectability does not apply to other robust (stable) learners, i.e. those that are less sensitive to perturbations and small changes in their training data. For example, Biggio et al. applied an evasion attack against a linear-kernel SVM trained on the MNIST dataset, which, while effective, generated attack images that were, visually and clearly, distinct from a legitimate character, and hence easy to detect as malicious \cite{biggio2012poisoning}. This is a stark difference from the evasion attacks against deep neural networks (relatively less robust classifiers) that have been shown to be very difficult to detect \cite{cao2017mitigating}. 

The research into defending deep learning models can be split into two primary approaches, detecting adversarial examples and building robust classifiers \cite{cao2017mitigating}. While there has been significant work into detecting adversarial examples \cite{Feinman2017, Gong2017, Hendrycks2017}, they have had limited success. Much of the research in defending deep learning models has gone into increasing the robustness of the model though either adversarial training \cite{Ororbia2016, Madry2017, Kurakin2017} or defensive distillation \cite{Papernot2016, Papernot2017}. However, it has been shown that even more robust machine learning algorithms are still vulnerable to adversarial examples \cite{biggio2012poisoning}, \cite{xiao2015feature}. In this paper, we provide evidence that although more robust algorithms are still vulnerable to adversarial examples, the adversarial examples are easier to detect.

\begin{figure}
\centering
\begin{tikzpicture}

\node[align=center] at (2.5,-0.5) {\textbf{Detectability}};
\node[align=center] at (0,-0.5) {Hard to detect \\ attack instances};
\node[align=center] at (5,-0.5) {Easy to detect \\ attack instances};

\node[align=center] at (-1,2) {\textbf{Instability}};
\node[align=center] at (-1,1) {Robust \\ Learners};
\node[align=center] at (-1,3) {Unstable \\ Learners};

\draw[dashed, line width=0.5mm, color=gray!30] (0,1) -- (1,0);
\draw[dashed, line width=0.5mm, color=gray!30] (0,2) -- (2,0);
\draw[dashed, line width=0.5mm, color=gray!30] (0,3) -- (3,0);
\draw[dashed, line width=0.5mm, color=gray!30] (0,4) -- (4,0);
\draw[dashed, line width=0.5mm, color=gray!30] (1,4) -- (5,0);
\draw[dashed, line width=0.5mm, color=gray!30] (2,4) -- (6,0);

\draw[->] (0,0) -- (6,0);
\draw[->] (0,0) -- (0,4);

\draw[->, line width=0.75mm] (2,1) -- (4,3);
\draw[->, line width=0.75mm] (1,2) -- (3,4);
\node[align=center, rotate=45] at (2.5,2.5) {Attack Strength / \\ Learner Vulnerability};

\end{tikzpicture}
\caption{A conceptual representation of the impact of the robustness of the model vs. the detectability of adversarial examples. Given two machine learning models where one is a more robust learner, to generate an attack with the same impact on the learner (i.e. attack strength), the attack against the more robust learner will be easier to detect.}
\label{fig:detectibility_instability}
\end{figure}
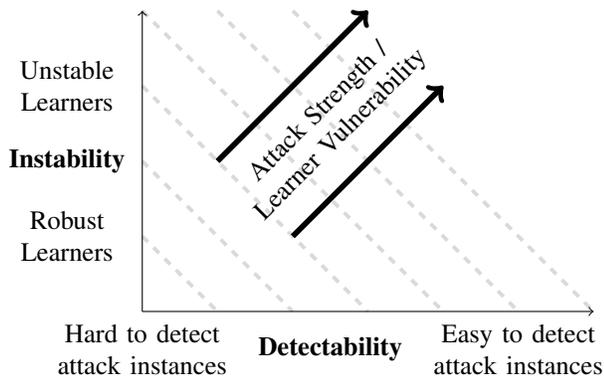

In order to secure a machine learning system, one must both have a robust model and some method of detecting adversarial examples. If the model is not robust, it may be difficult to detect adversarial examples. If the model is robust, one must still detect and remove the adversarial examples. These trends in the literature can be summarized by the detectability-instability diagram in Figure \ref{fig:detectibility_instability}.


Additionally, while adversarial machine learning is often defined as the intersection between machine learning and cybersecurity, very few connections have been made to cybersecurity. A common tool used to secure networks is an intrusion detection system that monitors network traffic for malicious activity \cite{roesch1999snort}; a similar tool does not presently exist for machine learning systems, and techniques such as novelty and outlier detection as well as other existing defense techniques (e.g. reject on negative impact (RONI) \cite{nelson2008roni}) should be combined in order to create something analogous to the cybersecurity intrusion detection systems.

\section{Conclusions and Future Work} \label{sec:conclusions}
\balance
In this work, we have shown that the attack proposed in \cite{xiao2015feature} can be easily defeated using outlier detection techniques. In response, we have proposed a modified attack that allows the attacker greater control over the strength of the attack in order to evade these detection techniques. Our results show a clear correlation between the attack strength and detectability of adversarial attack instances when attacking LASSO regression augmented with novelty and outlier detection.

Future work includes testing the improved attack algorithm using more sophisticated outlier detection terms, testing different combinations of defender-outlier detection methods and attacker outlier terms to better understand how much information the attacker needs about the defenders countermeasures, testing the use of multiple stages of countermeasures, and creating a framework for combining multiple attacks in order to poison black-box classifiers using existing perfect- and limited-knowledge attacks.

\section*{Acknowledgment}
This material is based upon work supported by the National Science Foundation under grant nos. 1310496 and 1429467. 

\bibliographystyle{IEEEtran}
\bibliography{ref}{}

\end{document}